**Chapter**

# ML Attack Models: Adversarial Attacks and Data Poisoning Attacks

*Jing Lin[1], Long Dang[2], Mohamed Rahouti[3], Kaiqi Xiong[1*]*

[1] ICNS Lab and Cyber Florida, University of South Florida, Tampa, FL jinglin@usf.edu, xiongk@usf.edu

[2] ICNS Lab & Computer Science and Engineering, University of South Florida, Tampa, FL longdang@usf.edu

[3] Computer and Information Sciences, Fordham University, Bronx, NY mrahouti@fordham.edu

* Corresponding Author

## Contents

# 4.1 INTRODUCTION

Machine learning (ML) is a research field focusing on the theory, properties, and performance of learning algorithms and systems. The ML field is considered highly interdisciplinary as it is built on the top of various disciplines, including, but not limited to, statistics, information theory, cognitive science, and mathematics (e.g., optimization theory). ML techniques have been intensively studied for many years. Notably, the achievement in many recent research studies related to smart city applications has been made possible using ML advances, where AI safety is directly relevant to ML security. As another example, ML implementations in an intelligent transportation system (ITS) can be deployed to analyze data generated by the different parts of the system (e.g., roads, number of passengers, and commute mode). The analysis of collected data is consequently used for future planning and decision-making within a transportation scheme [1].

Moreover, deep learning (DL) is another vital component of the ML field. Unlike many traditional learning techniques based on shallow-structured learning paradigms, DL is based upon deep architectures, where unsupervised and/or supervised learning strategies are mainly used to learn hierarchical representations autonomously. DL architectures (deep architectures) are often enabled to capture more hierarchically launched statistical patterns (complicated inputs patterns) in comparison to shallow-structured learning architectures [2].

Many state-of-the-art ML models have outperformed humans in various tasks such as image classification [3]. With such outstanding performance, ML models are widely used today. However, the existence of adversarial attacks and data poisoning attacks really questions the robustness of ML models. For instance, Engstrom *et al.* [4] demonstrated that state-of-the-art image classifiers could be easily fooled by a small rotation on an arbitrary image. As ML systems are being increasingly integrated into safety and security-sensitive applications [5], adversarial attacks and data poisoning attacks pose a considerable threat. This chapter focuses on the two broad and important areas of ML security: **adversarial attacks** and **data poisoning attacks**.

In **adversarial attacks**, attackers attempt to perturb a data point $x$ to an adversarial data point $x'$ so that $x'$ is misclassified by an ML model with high confidence, although $x'$ is visually indistinguishable from its original data point $x$ by humans. A visual illustration of an adversarial attack is shown in Figure 4.1, which presents an $L_\infty$-norm-based fast gradient sign method (FGSM) attack [6]. Adversarial attacks can be conducted in different application domains, such as audio [7], text [8], network signals [9], and images [10]. They can be further classified based upon the knowledge level that adversaries maintain about a target model. The attack model here can be specified as either *black-box*, *white-box*, or *gray-box*. *Black-box* attacks refer to the case when an attacker has no access to the trained ML model. Instead, the attacker only knows the model outputs (logit, confident score, or label). Black-box attacks are commonly observed in online learning techniques used in anomaly detection systems that effectively retrain the detector when new data is available. In a *gray-box* attack, attackers may have some incomplete knowledge about the overall structure of a target model. Last, in a *white-box* threat model, attackers have complete knowledge about a target model, which therefore facilitates the task of generating adversarial examples and crafting poisoned input data. Even though there are more white-box attacks among the three types of attacks, black-box attacks are more practical attacks on ML systems because essential model information is often confidential and protected from a normal user interacting with a system. Furthermore, due to the phenomenon of attack transferability, white-box attacks can be utilized to attack black-box models (transfer attacks). In this chapter, we focus on the two primary threat models: white-box and black-box attacks.

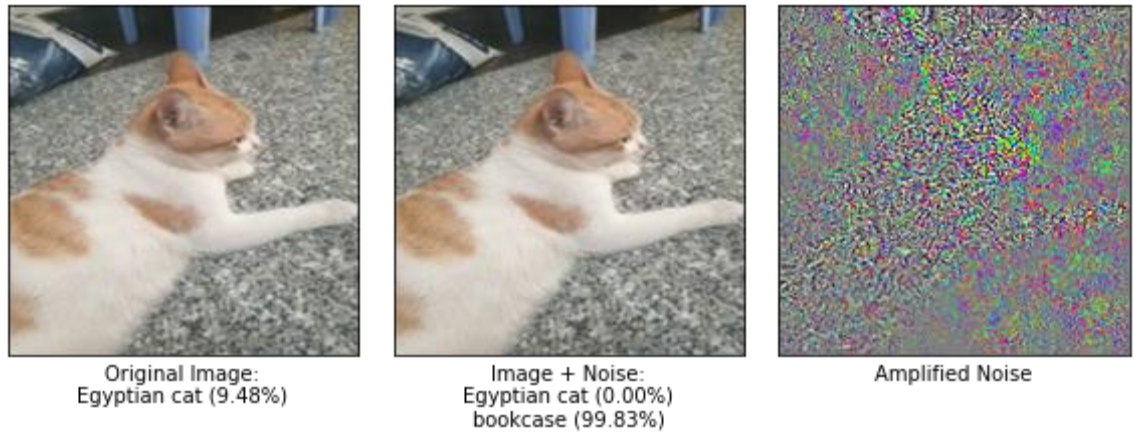

Figure 4.1: An example of $L_\infty$-norm-based FGSM attacks using a perturbation magnitude $\epsilon = 3$. Left: a base image (classified as a cat correctly by an inception v3 network [11]). Center: an adversarial image generated by the FGSM attacks (mislabeled as a bookcase with high confidence of 99.83%). Right: an (amplified) adversarial noise.

In **data poisoning attacks**, adversaries try to manipulate training data in an attempt

- to decrease the overall performance (i.e., accuracy) of an ML model,
- to induce misclassification to a specific test sample or a subset of the test sample, or
- to increase training time.

If an attacker has a specified target label to which a specific test sample is misclassified, the attack is called a targeted data poisoning attack; otherwise, it is called an untargeted data poisoning attack. Adversaries are assumed to either be able to contribute to the training data or have control over the training data. For instance, crowdsourcing is regarded as a vital data source for various smart cities, such as intelligent transportation systems and traffic monitoring [12]. Hence, such systems are vulnerable to data poisoning threats [13, 14] as data poisoning attacks are often difficult to be detected [15, 16].

This chapter will present adversarial attacks and data poisoning attacks in both white-box and black-box settings. These security attack classes are comprehensively discussed from different adversarial capabilities. The main objective of this chapter is to help the research community gain insights and implications of existing adversarial attacks and data poisoning attacks and increase awareness of potential adversarial threats when researchers develop future learning algorithms. The rest of this chapter is organized as follows. In section 4.2, we first provide some background information on support vector machines (SVMs) and neural networks. Then, we introduce some important definitions and typical white-box adversarial attacks in section 4.3. In section 4.4, we further discuss black-box adversarial attacks. Moreover, section 4.5 covers data poisoning attacks, along with their adversarial aspects. Finally, section 4.6 provides concluding remarks.

## 4.2 BACKGROUND

In this section, we introduce our notation before providing some background information on the SVMs and neural networks.

## 4.2.1 Notation

Here is a list of notations used throughout the chapter.

| Symbol | Meaning | Symbol | Meaning |
|---|---|---|---|
| $Z_+$ | Set of positive integers | $C$ | Classifier whose output is a label |
| $R$ | Set of real numbers | $\nabla$ | Gradient |
| $K$ | Number of classes | $\|\|.\|\|_p$ | $L_p$ -norm |
| $x$ | Original (clean, unmodified) input | $\delta$ | Perturbation $\delta = x' - x$ |
| $x'$ | Adversarial image (perturbed input image) | $\varepsilon$ | Some application-specific maximum perturbation |
| $y$ | Input $x$'s ground truth label | $g_i$ | The discriminant function for class $i$ |
| $t$ | Label of the targeted class in an adversarial attack | $f$ | Neural network (NN) model whose output is a confidence or probability score vector: $f(x) \in R^K$ |
| $L$ | The loss function of a NN | $Z$ | Logits |
| $\Theta$ | Parameters of $f$ | | |

## 4.2.2 Support Vector Machines

SVMs, known as large margin classifiers, are supervised learning models for both regression and classification problems. Formally, SVMs are ML methods to construct a hyperplane such that the distance between the hyperplane and the nearest training sample is maximized.

As an example, we focus on the hard-margin SVM for a linearly separable binary classification problem here. Given a training dataset of $n$ points written as $(x_1, y_1), \dots, (x_n, y_n)$, where $y_i$ is the class to which $x_i$ belongs. Each data point $x_i$ is a $p$-dimensional real vector and $y_i \in \{-1, 1\}$. The goal of SVMs is to search for the maximum-margin hyperplane that separates the group of points $x_i$ for which $y_i = 1$ from the group of points $x_i$ for which $y_i = -1$. The SVM, proposed by Cortes and Vapnik [17], finds the maximum margin hyperplane by solving the following quadratic programming (QP) problem:

$$\frac{1}{2}\|w\|^2, \qquad (4.2.1)$$

subject to the constraints, $y_i(w^T x_i + b) \geq 1$, for all $i = 1, 2, \dots, n$, where $w$ is the normal vector of the maximum margin hyperplane $y = w^T x + b$. This QP problem can be reformatted as an unconstrained problem using Lagrange multipliers $\alpha_i \geq 0$, for $i = 1, 2, \dots, n$, as follows.

$$L_p = \frac{1}{2}\|w\|^2 - \sum_{i=1}^n \alpha_i[y_i(w^T x_i + b) - 1]. \qquad (4.2.2)$$

Taking the derivative of the above expression with respect to $w$, we have

$$w = \sum_{i=1}^n \alpha_i x_i y_i. \qquad (4.2.3)$$

This indicates that the normal vector $w$ of the hyperplane depends on $\alpha_i, x_i$, and $y_i$. Similarly, taking the derivative of the above expression with respect to $w_0$, we obtain

Submitted as a book chapter in *AI, Machine Learning and Deep Learning: A Security Perspective,* CRC Press (Taylor & Francis Group), 2021.

$$\sum_{i=1}^{n} \alpha_i y_i = 0 \qquad (4.2.4)$$

Substituting equations (4.2.2) and (4.2.3) to $L_p$, we maximize the resulting $L_p$ with respect to $\alpha_i$ subject to equation (4.2.4) by applying a quadratic optimization method. Then, we can easily solve for $\alpha_i$. Once solved, most $\alpha_i$ are equal to zeros. The set of data points $(x_i, y_i)$ for which the corresponding $\alpha_i \neq 0$ are called support vectors. Since the normal vector $w = \sum_{i=1}^{n} \alpha_i x_i y_i$, the resulting hyperplane depends on the support vectors only.

However, if two classes are non-linear separable, a soft-margin SVM with a kernel function can be utilized to learn a non-linear decision boundary (linear in transformed space).

## 4.2.3 Neural Networks

In this subsection, we discuss neural networks, focusing on image-recognition systems since most existing adversarial examples introduced in the literature are related to image recognition.

Let $X = R^{hwc}$ be an input image space, where $h \in Z_+, w \in Z_+$, and $c \in Z_+$. For instance, a gray-scale image $x \in X$ has color channel $c = 1$, and a colored RGB image $x \in X$ has color channel $c = 3$. The element $x_{i,j,k}$ of vector $x$ denotes the value of pixel located at $(i, j, k)$. Each pixel value indicates how bright a pixel is, and the pixel value is an integer between 0 and 255. However, the pixel value is usually rescaled to be in the range of [0, 1].

A feed-forward neural network can be written as a function $f: X \rightarrow R^K$ that maps an input image $x \in X$ to an output $y \in R^K$, where $K \in Z_+$ is the number of classes, and $R$ is a set of real numbers. Each element $y_j$ of $y$ can be considered as the probability or likelihood that input $x$ belongs to class $j$, where $j$ is an integer between 1 and $K$. Each element $y_j$ of $y$ satisfies $y_j \in [0,1]$ and $\sum_{j=1}^{K} y_j = 1$. The output vector $y$ can be considered as a probability distribution over the discrete label set $\{1, 2, \ldots, K\}$. Furthermore, let $C: X \rightarrow \{1, 2, \ldots, K\}$ be the corresponding classifier that maps an input image $x \in X$ to $\{1, 2, \ldots, K\}$. The relationship between $C$ and the output vector $y$ is as follows:

$$C(x) = \underset{i=1,2,\ldots,K}{\mathrm{argmax}}\ y_i. \qquad (4.2.5)$$

A feed-forward neural network $f$ with $L \in Z_+$ layers can be considered as a composition function such that for each $x \in X$,

$$f(x) = f_L\left(f_{L-1}\left(f_{L-2} \ldots \left(f_1(x)\right)\right)\right), \qquad (4.2.6)$$

where each component function $f_i(x_i) = \sigma_i(\theta_i \cdot x_i + b_i)$, $\sigma_i$ is an activation function at layer $i$, $\theta_i$ is a matrix of weight parameters at layer $i$, and $b_i$ is a bias unit at layer $i$. Both $\theta = (\theta_1, \theta_2, \ldots, \theta_L)$ and $b = (b_1, b_2, \ldots, b_L)$ form model parameters $\Theta = (\theta, b)$. To keep things simple, we do not explicitly state the above function dependence on $\Theta$. If an attack model explicitly depends on it, we would write it as $f(x|\Theta) = y$. To find model parameters $\Theta$ such that the classifier $C$ can predict most instances in $X$ correctly, a stochastic gradient descent method is often utilized to minimize a loss function $L: X \times \{1, 2, \ldots, K\} \times P \rightarrow R^+$, where $P$ is parameter space. To keep things simple, we sometimes simply use $L: X \times \{1, 2, \ldots, K\} \rightarrow R^+$ when there is no confusion.

Let us further discuss activation function $\sigma_i$ as it is essential in expanding a linear hypothesis space to a non-linear hypothesis space. Rectified Linear Unit ($ReLU$) is widely used in ML/DL applications, especially in an image classification task. $ReLU: R \rightarrow R$ is defined on a real number set $R$ into a real number set $R$ such that $ReLU(x) = \max(0, x)$ for each $x \in R$. Therefore, the $ReLU$ is differentiable at all the points except $x = 0$. Usually, data scientists are interested in a non-linear activation function that is bounded and has a smooth gradient. Boundedness is an important property in preventing the return values of the activation function from becoming overly large. A smooth gradient is desirable since it makes back-propagation possible. However, $ReLU$ is neither upper-bounded nor non-differentiable at $x = 0$. Thus, why is it so popular and useful compared to traditional non-linear activation functions (such as sigmoid and hyperbolic tangent functions) that are bounded and with a smooth gradient? Existing studies have not laid a theoretical foundation to answer this question yet. However, the only non-differentiable point for $ReLU$ is when $x = 0$ and an input value of exactly zero for a $ReLU$ activation is not likely at any stage of the calculation, so back-propagation is working, although $ReLU$ is not everywhere differentiable.

There are many variations of $ReLU$ developed for the past few years, such as the Exponential Linear Unit (Exponential $LU$ or simply $ELU$), the Gaussian Error Linear Unit ($GELU$), and Leaky $ReLU$. For instance, $ELU$ transforms the negative input through an exponential function instead of assigning all negative input to be zero. GELU [18], an empirical improvement of $RELU$ and $ELU$ activations, is defined as follows:

$$\text{GELU}(x) = \left(\frac{x}{2}\right)\left[1 + \tanh\left(\sqrt{\frac{2}{\pi}}(x + 0.044715x^3)\right)\right]. \quad (4.2.7)$$

GELU has been the state-of-the-art activation function in Natural Language Processing for the past few years. Moreover, the Leaky ReLU [19] scales down the negative input value through multiplying it by a small positive constant $\alpha \in R_+$, e.g., α= 0.01 instead of returning zero when $x < 0$.

Lastly, we discuss the softmax function, a popular activation function for the last layer of a multi-classification model. The softmax function is a generalization of the logistic function or sigmoid function, and can be defined as follows: $Softmax: R^K \to [0,1]^K$ such that the $i^{th}$ element of the output for an input $z \in R^K$ is

$$Softmax_i(z) = \frac{exp\,(z_i)}{\sum_{j=1}^{K} exp\,(z_j)}. \quad (4.2.8)$$

## 4.3 WHITE-BOX ADVERSARIAL ATTACKS

We start with a reasonably comprehensive definition of adversarial attacks based on $L_p$-norm.

> **Definition 1 (Adversarial Attack):** Let $x \in R^n$ be a legitimate input data that is correctly classified as class $y$ by an ML classifier $f$. Given a target class $t$ such that $t \neq y$. An adversarial attack is a mapping $\alpha: R^n \to R^n$ such that the adversarial example $\alpha(x) = x'$ is misclassified as class $t$ by $f$, whereas the difference between $x'$ and $x$ is trivial, i.e., $||x' - x||_p < \epsilon$ for some small value $\epsilon$.

The $L_p$-norm $||x' - x||_p$ used in Definition 1 measures the magnitude of perturbation generated by an adversarial attack $\alpha$. The $L_p$-norm of a vector $v$ is defined as

$$||v||_p = (\sum_{i=1}^{n} |v_i|^p)^{\frac{1}{p}}, \quad (4.2.10)$$

where $v_i$ is the $i^{th}$ component of $v$. Here are three frequently used norms in literature.

- $L_0$-norm:

$$||x' - x||_0 = \sum_{i=1}^{n} J[x' \neq x], \text{ where } J[x' \neq x] = \begin{cases} 1, & x' \neq x \\ 0, & otherwise \end{cases},$$

  counts the number of coordinates $i$ such that $x_i \neq x_i'$, where $x_i$ and $x_i'$ are the $i^{th}$ component of $x$ and $x'$, respectively. The norm is useful when an attacker wants to limit the number of attack pixels without limiting the size of the change to each pixel.

- $L_2$-norm:

$$\|x' - x\|_2 = \sqrt{\sum_{i=1}^{n}(x'_i - x_i)^2}$$

  measures the Euclidean distance between $x'$ and $x$.

- $L_\infty$-norm or the Chebyshev distance:

$$\|x' - x\|_\infty = \max(|x_1' - x_1|, |x_2' - x_2|, \ldots, |x_n' - x_n|)$$

  measures the maximum absolute change to any of the coordinates of $x$.

Definition 1 is recognized as "targeted adversarial attacks." The "targeted" attacks comprise a target class $t$ and a function $\alpha$ aims at identifying a legitimate input data $x'$ (called adversarial example) such that the classifier $f$ misclassifies $x'$ as an instance of class $t$. In some scenarios, an attacker may not have a target class on her/his mind. Their goal is simply to mislead the classifier $f$ to misclassify an adversarial example $x'$ to a class other than its original class $y$.

This section mainly focuses on white-box attacks, assuming that an attacker has complete knowledge about the trained ML model. When attackers have access to this essential information, they can modify any arbitrary clean input using adversarial attack models. White-box attacks are the worst-case scenario because attackers have access to a model architecture and model weights.

## 4.3.1 L-BGFS Attack

Adversarial examples are often generated based on either first-order gradient information (such as the FGSM and DeepFool attack [20]) or gradient approximations. We first discuss some classical gradient-based attacks. In this case, the goal of an attacker is either to minimize the norm of the added perturbation that is needed to cause misclassification or to maximize the loss function model $f$ with respect to an input data.

*Szegedy et al.* [21] demonstrate the first targeted adversarial attack on deep neural networks. They find a minimal distorted adversarial example $x'$ by solving the following optimization problem:

$$\min_{x'} \|x' - x\|_p, \tag{4.3.1}$$

subject to the constraints, $C(x') = t$ and $x' \in [0, 1]^n$, where $\|.\|_p$ is the $L_p$-norm. Finding a precise solution to this problem is very challenging since the constraint $C(x') = t$ is non-linear. *Szegedy et al.* replaced the constraint $C(x') = t$ with the continuous loss function $L$, such as the cross-entropy for classification. That is, they solved the following optimization problem instead:

$$\min_{x'} c \|x' - x\|_p + L(x', t), \tag{4.3.2}$$

subject to the constraint $x' \in [0, 1]^n$, where constant $c > 0$. By first adaptively choosing different values of constant $c$ and then iteratively solving the problem (4.3.2) using the L-BFGS algorithm [22] for each selected $c$, an attacker can obtain an adversarial example $x'$ that has the minimal distortion to $x$ and the $C(x') = t$. The L-BGFS attack is formally defined as follows.

**Definition 3 (L-BGFS Attack by *Szegedy et al.* 2014):** Let $x \in R^n$ be a legitimate input data that are correctly classified as class $y$ by an ML classifier $C$. Given a loss function $L$, the L-BFGS attack generates an adversarial example $x' \in R^n$ by minimizing the following objective function:

$$c \| x' - x \|_p + L(x', t),$$

subject to the constraint $x' \in [0, 1]^n$, where $x'$ is misclassified as class $t \neq y$ by $C$.

## 4.3.2 Fast Gradient Sign Method

Unlike the L-BFGS attack, the fast gradient sign method (FGSM) proposed by *Goodfellow et al.* [6] focuses on finding an adversarial perturbation limited by $L_\infty$-norm efficiently rather than producing an optimal adversarial example. In the training of an ML model, a given loss function is minimized to find an optimal parameter set $\Theta$ for a classifier $C$ so that $C$ classifies most training data correctly. In contrast, FGSM maximizes the loss function as it tries to make the classifier perform poorly on the adversarial example $x'$. Therefore, the FGSM perturbed image for an untargeted attack is constructed by solving the following maximization problem:

$$\max_{x' \in [0,1]^n} L(x', y), \tag{4.3.3}$$

subject to the constraint $\| x' - x \|_\infty \leq \varepsilon$, where $y$ is the ground-truth label of the original image $x$, and $\varepsilon$ is some application-specific maximum perturbation.

Applying the first-order Taylor series approximation, we have

$$L(x', y) = L(x + \delta, y) \approx L(x, y) + \nabla_x L(x, y)^T \cdot \delta, \tag{4.3.4}$$

where $\delta = x' - x$ is an adversarial perturbation, and $\nabla_x L(x, y)^T$ refers to the gradient of the loss function $L$ with respect to input $x$ and it can be computed quickly by a back-propagation algorithm.

The objective function in (4.3.3) is rewritten as:

$$\min_{\delta} \ -L(x, y) - \nabla_x L(x, y)^T \cdot \delta, \tag{4.3.5}$$

subject to the constraint $\| \delta \|_\infty \leq \varepsilon$, where we convert the maximization problem to the minimization problem by flipping the sign of the two components. Note that $L(x, y)$ and $\nabla_x L(x, y)^T$ are constant and already known because the attack is in the white-box attack setting.

Furthermore, since $0 \leq \| \delta \|_\infty \leq \varepsilon$, we have

$$\delta = \varepsilon \cdot sign(\nabla_x L(x, y)), \tag{4.3.6}$$

where $sign$ function outputs the sign of its input value.

For the targeted attack setting, an attacker tries to minimize the loss function $L(x, t)$, where $t$ is the target class. In this case, we can show that

$$\delta = -\varepsilon \cdot sign(\nabla_x L(x, t)). \tag{4.3.7}$$

Formally, we can define FGSM with $L_\infty$-norm bounded perturbation magnitude as follows.

**Definition 4 (FGSM Adversarial Attack by *Goodfellow et al.* 2014):** Let $x \in R^n$ be a legitimate input data that are correctly classified as class $y$ by an ML classifier $f$. The FGSM with $L_\infty$-norm bounded perturbation magnitude generates an adversarial example $x' \in R^n$ by maximizing the loss function $L(x', y)$ subject to the constraint $\| x' - x \|_\infty \leq \varepsilon$. That is,

$$x' = \begin{cases} x + \varepsilon \cdot sign(\nabla_x L(x, y)^T), untargeted \\ x - \varepsilon \cdot sign(\nabla_x L(x, t)^T), targeted \ on \ t. \end{cases}$$

Furthermore, we can easily generalize the attack to other $L_p$-norm attacks[1]. For example, we can use the Cauchy-Schwarz inequality, a particular case of the Holder's inequality with $p = q = 2$, to find the lower bound of the objective function in problem (4.3.5) subject to the constraint $\| x' - x \|_2 \leq \varepsilon$ and to obtain the perturbation for an untargeted FGSM for $L_2$-norm-based constraint

$$\delta = \frac{\varepsilon \nabla_x L(x,y)}{\|\nabla_x L(x,y)\|_2}. \tag{4.3.8}$$

Unless otherwise specified, we will focus on the attacks with the $L_\infty$-norm bounded perturbation magnitude in the rest of this chapter.

## 4.3.3 Basic Iterative Method

The basic iterative method proposed by *Kurakin et al.* [23] is an iterative refinement of the FGSM. BIM uses an iterative linearization of the loss function rather than the one-shot linearization in FGSM. Furthermore, multiple smaller but fixed step sizes $\alpha$ are taken instead of a single larger step size $\varepsilon$ in the direction of the gradient sign. For instance, in an untargeted attack with $L_\infty$-norm bounded perturbation magnitude, the initial point $x^{(0)}$ is set to the original instance $x$, and in each iteration, the perturbation $\delta^{(k)}$ and $x^{(k+1)}$ are calculated as follows:

$$\delta^{(k)} = \alpha \times sign\left(\nabla_x L(x^{(k)}, y)\right) \tag{4.3.9}$$

$$x^{(k+1)} = Clip_{x,\varepsilon}\{x^{(k)} + \delta^{(k)}\}, \tag{4.3.10}$$

where $\varepsilon > 0$ denotes the size of the attack and $Clip_{x,\varepsilon}(x')$ is a projection operator that projects the value of $x'$ onto the intersection of the box constraint of $x$ (for instance, if $x$ is an image, then a box constraint of $x$ can be the set of integer values between 0 and 255) and the $\epsilon$ neighbor ball of $x$. The above procedure ensures that the produced adversarial examples are within $\epsilon$ bound of the input $x$. Using a reduced perturbation magnitude $\alpha$ limits the number of active attack pixels and thus, prevents a simple outlier detector from detecting the adversarial examples.

> **Definition 5 (BIM Adversarial Attack by *Kurakin et al.* 2016):** The BIM attack generates an $L_\infty$-norm bounded adversarial example $x' \in R^n$ by iteratively maximizing the loss function $L(x', y)$, subject to the constraint $\| x' - x \|_\infty \leq \varepsilon$. For iteration $k$, $x^{(k+1)}$ is calculated as follows:
>
> $$\delta^{(k)} = \alpha \times sign\left(\nabla_x L(x^{(k)}, y)\right)$$
>
> $$x^{(k+1)} = Clip_{x,\varepsilon}\{x^{(k)} + \delta^{(k)}\},$$
>
> where $x^{(0)} = x, \alpha < \varepsilon$ is a smaller but fixed perturbation magnitude in each iteration, and the number of iterations is chosen by the user.

A popular variant of BIM is the projected gradient descent (PGD) attack, which uses a uniformly random noise as an initialization instead of setting $x^{(0)} = x$ [24]. The random initialization is used to explore the input space. Later, Croce and Hein [24] improved PGD by adding a momentum term for a gradient update and utilizing the exploration vs. exploitation concept for optimization. They called this approach the auto projected gradient descent (Auto-PGD) and showed that Auto-PGD is more effective than PGD [24]. It is further improved with AutoAttack [24], an ensemble of various attacks. Indeed, AutoAttack combines three white-box attacks with one black-box attack. The four attack components are two extensions of Auto-PGD attacks (one maximizes cross-entropy loss, and another maximizes the difference of logistic ratio loss) and two existing supporting attacks, a fast adaptive boundary (FAB) attack [25] and a square attack [26], respectively. It has been shown that using the ensemble can improve the attack effectiveness over multiple defense strategies [25]. An attack is considered successful for each test example if at least one of the four attack methods finds an adversarial example.

---

[1] Lecture note: Chan, Stanley H. "Adversarial Attack." Machine Learning ECE595, April 2019, Purdue University. Portable document format (PDF). [23]

Furthermore, Auto-Attack can run entirely without predefined user inputs across benchmark datasets, models, and norms. Therefore, it can provide a reliable, quick, and parameter-free evaluation tool when a researcher develops adversarial defenses [24]. We will introduce fast adaptive boundary attacks in section 4.3.5 and square attacks in section 4.4.4.

## 4.3.4 DeepFool

The DeepFool [20] attack is an untargeted white-box attack. Like *Szegedy et al.* [21], *Moosavi-Dezfooli et al.* [20] studied the minimally distorted adversarial example problem. However, rather than finding the gradient of a loss function, DeepFool searches for the shortest distance from the original input to the nearest decision boundary using an iterative linear approximation of the decision boundary/hyperplane and the orthogonal projection of the input onto the approximated decision boundary.

*Moosavi-Dezfooli et al.* searched the shortest distance path for an input $x$ to cross the decision boundary and get misclassified. Formally, the DeepFool attack can be defined as follows.

> **Definition 6 (DeepFool Adversarial Attack by *Moosavi-Dezfooli et al.* 2016):** The DeepFool $L_2$-norm attack generates adversarial example $x' \in R^n$ by solving the following optimization problem:
> 
> $$\min_{x'} \;\; \|x' - x\|_2^2, \qquad (4.3.11)$$
> 
> subject to the constraint $x' \in g$, where $g$ is the decision boundary that separating class $c_i$ from class $c_j$ ($j \neq i$).

To find such a path, an attacker first approximates the decision boundary using the iterative linear approximation method and then calculates the orthogonal vector from $x$ to that linearized decision boundary. Furthermore, we present the solution to the binary classification case and extend the solution to the multi-classification case later.

Given a non-linear binary classifier with discriminant functions $g_1$ and $g_2$, the decision boundary between class 1 and class 2 is $g = \{x: g_1(x) - g_2(x) = 0\}$. A per-iteration approximate solution $x'_{i+1}$ can be derived by approximating the decision boundary/hyperplane $g$ based on the first-order Taylor expansion:

$$g(x') \approx g(x_i) + \nabla_x g(x_i)^T (x' - x_i), \qquad (4.3.12)$$

where $x_i$ is the $i^{\text{th}}$ iterate of the solution and $x_0 = x$. Then, we solve the following optimization problem:

$$\min_{x'} \;\; \|x'_{i+1} - x_i\|, \qquad (4.3.13)$$

subject to $g(x_i) + \nabla_x g(x_i)^T (x' - x_i) = 0$ by finding the saddle node of the Lagrangian given by:

$$\tfrac{1}{2} \|x'_{i+1} - x_i\|_2^2 + \lambda(g(x_i) + \nabla_x g(x_i)^T (x' - x_i)). \qquad (4.3.14)$$

The optimization problem can be solved to obtain the saddle node

$$x_{i+1} = x_i - \left(\frac{g(x_i)}{\|\nabla_x g(x_i)^T\|_2}\right) \frac{\nabla_x g(x_i)}{\|\nabla_x g(x_i)\|_2}.$$

This saddle node can be considered as the orthogonal projection of $x_i$ onto the decision hyperplane $g$ at iteration $i$, and the iteration stops when $x_{i+1}$ is misclassified; i.e., $sign(g(x_{i+1})) \neq sign(g(x_0))$.

The multi-class (in one-vs-all scheme) approach of the DeepFool attack follows the iterative linearization procedure in the binary case, except that an attacker needs to determine the closest decision boundary $l$ to the input $x$. The iterative linear approximation method used to find the minimum perturbation is a greedy method that may not guarantee the global minimum. Moreover, the closest distance from the original input to the nearest decision boundary may not be equivalent to the minimum difference observed by human eyes. In practice, DeepFool usually generates small unnoticeable perturbations.

So far, we have only considered untargeted DeepFool attacks. A targeted version of DeepFool can be achieved if the input $x$ is pushed towards the boundary of a target class $t \neq y$ subject to $C(x'_{i+1}) = t$. Furthermore, the DeepFool attack can also be adapted to find the minimal distorted perturbation for any $L_p$-norm, where $p \in [1, \infty]$. If interested, see [28] for details.

## 4.3.5 Fast Adaptive Boundary Attack

The fast adaptive boundary attack, proposed by Croce and Hein [25], is an extension of the DeepFool attack [20]. Specifically, the computation of $x_i'$ for a FAB attack follows the DeepFool algorithm closely. The main differences between the two types of attacks are that a DeepFool attacker projects $x_i'$ onto the decision hyperplane $g$ only, $Proj_g(x_i')$, whereas a FAB attacker projects $x_i'$ onto the intersection of $[0,1]^n$ and the approximated decision hyperplane $g$, $Proj_{g \cap [0,1]^n}(x_i')$, with the following additional steps.

- Add a momentum term $\alpha$ to regulate the influence of the additional term $x + \eta * Proj_{g \cap [0,1]^n}(x)$ on the modified image $x_i' + \eta * Proj_{g \cap [0,1]^n}(x_i')$. Thus, $x_i'$ is biassed toward $x$ and hence, the $x_i'$ generated by the FAB attack method is closer to the input $x$ than one generated by the DeepFool attack. The momentum term $\alpha$ is updated for each iteration $i$ as follows. First, a FAB attacker computes the minimal perturbations $\delta_i = Proj_{g \cap [0,1]^n}(x_i')$ and $\delta_0 = Proj_{g \cap [0,1]^n}(x_0)$ as the current best point $x_i'$ and the input $x$ are projected onto the intersection of $[0,1]^n$ and the approximated linear decision boundary, respectively. Then, $\alpha$ is chosen as follows.

$$\alpha = \min\left\{\frac{\left|Proj_{g \cap [0,1]^n}(x_i')\right|}{\left|Proj_{g \cap [0,1]^n}(x_i')\right| + \left|Proj_{g \cap [0,1]^n}(x)\right|}, \alpha_{\max}\right\} \in [0, 1], \quad (4.3.15)$$

where $\alpha_{max} \in [0,1]$ is a hyperparameter that provides an upper limit value for $\alpha$. $\alpha$ is defined such a way that the per-iteration $x_i'$ might go too further away from $x$.

- Add a backward step at the end of each iteration to increase the quality of the adversarial example even further. The DeepFool stops when $x_i'$ is successfully misclassified. However, the FAB attack has this additional step for enhancement. The backward step is simply a movement of $x_i'$ closer to the input $x$ by calculating the linear combination of $x$ and $x_i'$ with a hyperparameter $\beta \in (0,1)$. Croce and Hein [25] used $\beta = 0.9$ for all their experiments.

$$x_i' = (1 - \beta)x + \beta x_i' = x + \beta(x_i' - x). \quad (4.3.16)$$

- Add random restarts to widen the search space for adversarial examples. That is, instead of initializing $x_0' = x$, a FAB attacker sets $x_0'$ as a random sample in a $\epsilon'$-neighborhood of $x$, where $\epsilon' = \min(||x_i' - x||_p, \epsilon)$.

- Add a final search step to further reduce the distance between the adversarial example and its original input. This step uses a modified binary search on $x_i'$ and $x$ to find a better adversarial example within a few iterations; e.g., Croce and Hein [25] set the number of iterations to 3 for their experiments. For details of the final search step and the random restarts, see [25].

## 4.3.6 Carlini and Wagner's Attack

Carlini and Wagner's (C&W) attack [27] aims to find the minimally disturbed perturbation, where Carlini and Wagner solved the box-constrained optimization problem defined in section 4.3.1. Specifically, they converted the box-constrained optimization problem into an unconstrained optimization problem, which can then be solved through standard optimization algorithms instead of the L-BFGS algorithm. Carlini and Wagner investigated three different methods to get rid of the box-constraint $x' \in [0,1]^n$, projected gradient descent, clipped gradient descent, and change of variables. They concluded that the method of the change of variables is most effective in generating adversarial examples fast. That is, they introduced a set of new variables $w_i \in R, i = 1, 2, \ldots, n$ such that

$$x'_i = \tfrac{1}{2}(\tanh(w_i) + 1). \tag{4.3.17}$$

Though $x'_i \in [0,1]$, $w_i$ can be any real number so that the box-constraint is eliminated.

Furthermore, the constraint $C(x') = t$ is highly non-linear. Carlini and Wagner [29] considered seven objective candidate functions $g$ to replace the constraint $C(x') = t$. Each $g$ satisfies the condition that $C(x') = t$ if and only if $g(x') \leq 0$ (for targeted attacks) and $C(x') \neq y$ if and only if $g(x') \leq 0$ (for untargeted attacks). For instance, C&W's $L_2$-norm attack has $g$ defined as follows.

$$g(x') = \begin{cases} \max\{\max_{j \neq t} Z_j(x') - Z_t(x'), -\kappa\}, & \text{target on } t \\ \max\{Z_y(x') - \max_{j \neq y} Z_j(x'), -\kappa\}, & \text{untargeted,} \end{cases} \tag{4.3.18}$$

where $Z$ is the logits and $Z_j$ is the $j^{th}$ element of the logits; $\kappa \geq 0$ is a parameter that controls the strength of adversarial examples. The default value is zero for experiments in [29]. However, increasing $\kappa$ can generate adversarial examples with a higher transfer success rate.

Instead, Carlini and Wagner [27] solved the optimization problem:

$$\min_{x'} \parallel x' - x \parallel_p + c \cdot g(x'), \tag{4.3.19}$$

where $c > 0$ is a regularization parameter that controls the relative importance of the $L_p$-norm perturbation over $g$. If a large value of $c$ is chosen, the attacker generates an adversarial example which is further away from the base example but misclassified with high confidence and vice versa. Carlini and Wagner [27] used a modified binary search to find the smallest value of $c$ for which the solution $x'$ to the optimization problem (4.3.19) satisfies the condition $g(x') \leq 0$. Alternative to the modified binary search, a line search can also be used to find the adversarial example with the minimal $L_p$ distance from $x$.

Furthermore, Carlini and Wagner considered three types of attacks, $L_0$-norm attack, $L_2$-norm, and $L_\infty$-norm attacks. The formal definition of the C&W attack is presented as follows.

> **Definition 7 (C&W Adversarial Attack by *Carlini et al.* 2017):** The C&W attack generates an adversarial example $x' \in R^n$ by solving the following optimization problem:
>
> $$\min_{x'} \parallel x' - x \parallel_2^2 + c \cdot g(x'),$$
>
> where $x'_i = \tfrac{1}{2}(tanh(w_i) + 1)$, $c > 0$ is the regularization term, and

$$g(x') = \begin{cases} \max\left\{\max\limits_{j \neq t} Z_j(x') - Z_t(x'), -\kappa\right\}, \text{target on } t \\ \max\left\{Z_y(x') - \max\limits_{j \neq y} Z_j(x'), -\kappa\right\}, \text{untargeted}. \end{cases}$$

Note that the standard gradient descent algorithm can be used to find the solution of the above minimization problem.

Carlini and Wagner [27] showed that the C&W attack is very powerful, and ten proposed defense strategies cannot withstand C&W attacks constructed by minimizing defense-specific loss functions [28]. Furthermore, the C&W attack can also be used to evaluate the efficacy of potential defense strategies since it is one of the strongest adversarial attacks [27].

In the next two subsections, we will discuss some adversarial attacks other than $L_p$-norm-based attacks.

## 4.3.7 Shadow Attack

In this subsection, we introduce a class of adversarial attacks called "semantic" adversarial attacks on image datasets that usually generate adversarial examples through a larger perturbation on an image in terms of $L_p$-norm but are still semantically similar to its original image. A typical example of the attack is simply a small color shift, rotation, shearing, scaling, or translation of a natural image to fool a state-of-the-art image classifier [4, 29]. The attack is successful due to differences in the way the ML model recognizes an image and the human visual recognition system. A semantic adversarial example can be formally defined as follows.

**Definition 8 (Semantic Adversarial Attack):** Let $x \in R^n$ be a legitimate input data that are correctly classified as class $y$ by an ML classifier. A semantic adversarial attack is a mapping $\alpha: R^n \to R^n$ such that the adversarial example $\alpha(x) = x'$ is misclassified as class $t \neq y$ by the classifier and $\beta(x) = \beta(x')$, where $\beta$ is the human visual recognition system.

The semantic adversarial attacks can be either white-box or black-box. While we discuss one of the white-box semantic adversarial attacks in detail here, subsection 4.4.3 will be concerned with a spatial transformation attack, a box-box semantic adversarial attack.

The shadow attack [30] is a type of "semantic" adversarial examples that is different from the attacks introduced in subsections 4.3.1- 4.3.6 as it targets not only the classifier's output label but also the certifiable defenses [31, 32]. This attack reveals that certifiable defenses are not inherently secure. Furthermore, shadow attacks can also be considered as a generalization of PGD attacks with three penalty terms that minimize the perceptibility of perturbations. Formally, shadow attacks are defined as follows, with three penalty terms added to the loss term.

**Definition 9 (Shadow Attack by *Ghiasi et al.* 2020):** The shadow attack generates the adversarial example by solving the following optimization problem:

$$\max_{\delta} L(x + \delta, \theta) - \lambda_{tv} TV(\delta) - \lambda_c C(\delta) - \lambda_s D(\delta), \quad (4.3.20)$$

where $\theta$ is the model weights, $x$ is an arbitrary natural image, $\delta$ is the perturbation added to image $x$, $L$ is a loss function, $\lambda_c > 0, \lambda_{tv} > 0$, and $\lambda_s > 0$ are scalar penalty weights, $TV(\delta)$ is the total variation of $\delta$, $C(\delta) = ||Avg(|\delta_R|)||_p + ||Avg(|\delta_B|)||_p + ||Avg(|\delta_G|)||_p$ restricts the change in the mean of each color channel, and $D(\delta) = ||\delta_R - \delta_B||_p + ||\delta_R - \delta_G||_p + ||\delta_G - \delta_B||_p$ promotes even perturbation between the color channels.

Note that the total variation $TV(\delta)$ is calculated element-wise as $TV(\delta_{i,j}) = anisotropic - TV(\delta_{i,j})^2$ [30], where $anisotropic - TV(\delta_{i,j})^2$ estimates the $L_1$-norm of the gradient of $\delta_{i,j}$.

## 4.3.8 Wasserstein Attack

Different from most adversarial attacks that focus on $L_p$-norm, the Wasserstein attack uses the Wasserstein distance (also known as the optimal transport, Kantorovich distance, or Earth mover's distance) to generate Wasserstein adversarial examples. $L_p$-norms, though used extensively in adversarial attacks, do not capture the image transforms such as rotation, translation, distortion, flipping, and reflection.

**Definition 10** (The Wasserstein distance between two inputs $x$ and $y$). Let $x \in R_+^n$ and $y \in R_+^n$ be two inputs that are normalized with $\sum_{i=1}^n x_i = 1$ and $\sum_{j=1}^n x_j = 1$. Furthermore, let $C \in R^{n \times n}$ be a cost matrix such that each element $C_{i,j}$ of $C$ measures the cost of moving mass from $x_i$ to $y_j$ (i.e., the Euclidean distance between the pixel $x_i$ and the pixel $y_j$, if $x$ and $y$ are images). Then, the Wasserstein distance between two inputs $x$ and $y$ is:

$$d_W(x, y) = <T, C>, \quad (4.3.21)$$

subject to $T1 = x, T^T 1 = y$, where $1 \in R^{n \times 1}$ is a column vector of ones, $T \in R^{n \times n}$ is a transport plan whose element $T_{i,j} > 0$ encodes how the mass moves from $x_i$ to $y_j$ and $<T, C>$ denotes the sum of the matrix-matrix element-wise multiplication. That is, $<T, C> = \sum_{i=1}^n \sum_{j=1}^n T_{ij} * C_{ij}$.

Probabilistically, you can think of a transport plan as a joint probability of $x$ and $y$ [33]. Although the Wasserstein distance has many theoretical advantages, calculating a Wasserstein distance between two images is computationally expensive as it requires solving a linear programming problem with $n \times n$ number of variables. *Wong et al.* [34] developed a fast approximate projection algorithm, called the projected Sinkhorn iteration, that modifies the standard PGD attack by projecting onto a Wasserstein ball. Formally, *Wong et al.* proposed to solve the following objective function as follows.

**Definition 11** (**The Wasserstein adversarial example by *Wong et al.* 2019**). The Wasserstein attack generates a Wasserstein adversarial example $x' \in R^n$ by minimizing the following objective function:

$$||x - x'||_2^2 + \frac{1}{\lambda}\sum_{ij} T_{ij} \log \log (T_{ij}), \quad (4.3.22)$$

subject to the Wasserstein ball constraints, $T1 = x, T^T 1 = x'$, and $d_W(x, y) < \epsilon$, where $\lambda \in R$.

Note that an entropy-regularized term, $\sum_{ij} T_{ij} \log \log (T_{ij})$, on the transport plan $T$, is included in the objective function so that projection is approximated; however, this entropy-regularized term speeds up the computation (near-linear time). To further reduce the computational cost, the local transport plan restricts the movement of each pixel to be within the $k \times k$ region of its original location. For instance, *Wong et al.* [34] set $k = 5$ for all their experiments.

## 4.4 BLACK-BOX ADVERSARIAL ATTACKS

Recent research on the adversarial attack has shown particular interest in black-box settings. Although many attacks require the attacker to have full access to the network architecture and weight, this information is often protected in commercialized products such as the vision system in modern self-driving vehicles. As a result, the black-box attack, in which an attacker has no knowledge of the network and the training set except the ability to query the probability/confidence score for each class (score-based black-box attack) or the label (decision-based black-box attack), is increasingly important to investigate. Furthermore, since black-box attack models do not require the knowledge of gradients as in most white-box attacks introduced in section 4.3, the defense strategies based on the masking gradient or non-differentiability do not work with the attacks introduced in this section.

In section 4.4.1, we introduce a transfer attack, a popular black box attack that utilizes the transferability of adversarial examples. This attack can even be applied to no-box settings, in which an attacker does not even have access to the output of a model. In section 4.4.2, we present two score-based black-box attacks, zeroth order optimization (ZOO) based attack and square attack. Then, we discuss three popular decision-based black-box attacks in section 4.4.3.

## 4.4.1 Transfer Attack

*Szegedy et al.* [6] highlight the transferability of adversarial examples through the generalization power of DNNs. That is, the same adversarial image can be misclassified by a variety of classifiers with different architectures or trained on different training data. This property is useful in black-box settings. When an attacker does not have knowledge of model architecture, they can generate a substitute model to imitate the target model and then apply white-box attack methods such as L-BGFS attack to the obtained substitute model in order to generate an adversarial example. This type of attack is called a transfer attack.

To do a transfer attack, an attacker needs to have a labeled training set first. With free query access, an attacker can feed the substitute training set to the target classifier in order to obtain the label for these instances. In the extreme case of limited query access, an attacker may have to do the labeling on their own or find a substitute training set that is already labeled. However, with the power of free query access, the generated model is usually more representative of the target model. With a labeled substitute training set, an attacker can select a substitute model that has a similar structure as the target model if they have any knowledge of the underlying structure. For instance, an image classifier usually has multiple CNN layers, whereas a sequential model may have some sort of RNN layers. Then, they can train the selected substitute model with the substitute training set. This trained substitute model serves as an attack surrogate. Finally, white box attack methods introduced in section 3.3 can be used to generate adversarial examples. In order to have a high transfer attack success rate, an FGSM, PGD, C&W, or other attack methods with high transferability are selected to generate adversarial examples. Furthermore, the adversarial perturbation required is usually somewhat larger in order to ensure the success of a transfer attack.

## 4.4.2 Score-based Black-box Attacks

In this section, we discuss two popular score-based attacks, ZOO and square attacks. The ZOO attack is to find adversarial examples based on zero-order-optimization, whereas the square attack generates adversarial examples by using derivative-free optimization (DFO). Below are the detailed discussions of the two attacks.

**ZOO Attack**

Different from the transfer attacks exploiting the transferability of adversarial images, *Chen et al.* [35] proposed a ZOO attack to directly approximate the gradients of the target model using confidence scores. Therefore, the ZOO attack is considered a score-based black-box attack, and it does not need to train a substitute model. The ZOO attack is as effective as the C&W attack and remarkably surpasses existing transfer attacks in terms of a success rate. *Chen et al.* [35] also proposes a general framework utilizing capable gradient-based white-box attacks for generating adversarial examples in the black-box setting.

The ZOO attack finds the adversarial example by also solving the optimization problem (4.3.19). Motivated by the attack loss functions (4.3.6.2) used in the C&W attack, a new hinge-like loss function [35] based on the log probability score vector $p = f(x')$ of the model $f$, instead of $Z$, is proposed as follows.

                                                                                                                                  

$$g(x') = \begin{cases} \max\left\{\max_{j \neq t}[f(x')]_j - [f(x')]_t, -\kappa\right\}, \text{target on } t \\ \max\left\{[f(x')]_y - \max_{j \neq y}[f(x')]_j, -\kappa\right\}, \text{untargeted}, \end{cases} \quad (4.4.1)$$

where $[f(x')]_j$ is the $j^{th}$ element of the probability score vector, and the parameter $\kappa \geq 0$ ensures a constant gap between the log probability score of the adversarial example classified as class $t$ and all remaining classes. The log probability score is used instead of a probability score since well-trained DNNs yield a significant high confidence score for a class compared to other classes and the log function lessens this dominance effect without affecting the order of confidence score. The ZOO attack is defined as follows.

**Definition 12 (ZOO Adversarial Attack by *Chen et al.* 2017):** A ZOO attack is a score-based black-box attack that generates the adversarial example $x'$ by solving the following optimization problem using the zeroth-order stochastic coordinate descent with a coordinate-wise ADAM.

$$\min_{x'} \; \| x' - x \|_p + c \cdot g(x')$$

subject to the constraint $x' \in [0,1]^n$.

In the white box setting, finding an adversarial example requires taking the gradient of the model function $\partial f(x)$. However, in the black-box setting, the gradient is inadmissible, and one can only use the function evaluation $f(x)$, which makes it a zeroth-order optimization problem. *Chen et al.* [35] provide a method for estimating the gradient information surrounding the victim sample $x$ by watching variations in prediction confidence $f(x)$ when the coordinate values of $x$ are adjusted [36]. *Chen et al.* [35] used the following symmetric difference quotient to approximate the gradient:

$$\frac{\partial f(x)}{\partial x_i} \approx \frac{f(x+he_i)-f(x-he_i)}{2h} \quad (4.4.2)$$

where $h$ is a small constant, and $e_i$ is a standard basis vector. For networks with a large input size $n$, e.g., Inception-v3 network [11] has $n = 299 \times 299 \times 3$, the number of model queries per gradient evaluate is $2 \times n$ (two function evaluation per coordinate-wise gradient estimation). This is very query inefficient.

To overcome this inefficiency, *Chen et al.* [29] proposed the five acceleration techniques.

1. Use zeroth-order stochastic coordinate descent with a coordinate-wise ADAM to minimize the objective function.

2. Evaluate the objective function in batches instead of one-by-one to take advantage of the GPU.

3. Reduce the attack space and perform the gradient estimation from a dimension reduced space $R^m$, where $m < n$, to improve computational efficiency. Specifically, a dimension reduction transformation $D: R^m \to R^n$ such as $D(\delta') = x' - x$, where $\delta' \in R^m$ is used to reduce the attack space. For instance, $D$ can be an up-scale operator that scales up an image, such as the bilinear interpolation method, discrete cosine transformation (DCT), or an autoencoder for attack dimension reduction [37]. Although this attack space dimension reduction method reduces the computation costs for high-resolution image datasets (such as ImageNet [38]), the reduced attack space also limits the ability to find a valid attack.

4. Use hierarchical attack to overcome the limited search space issue. Instead of just one transformation, a series of transformations $D_1, D_2, \ldots, D_v$ with dimensions $m_1, m_2, \ldots, m_v$, where $m_1 < m_2 < m_v$ are used to find the adversarial example. Start with the small attack space $m_1$, an attacker tries to find an adversarial example. If they are unsuccessful, then they increase the attack space to $m_2$. Otherwise, the process stops.

5. Choose which coordinates to update based on their "importance." For instance, the pixel near the

edge or corner of an image is less important than a pixel near the main object (usually near center) for an image classifier. *Chen et al.* [35] divided an image into $8 \times 8$ regions and defined the "importance" as the absolute values of pixel value changes in each region.

Note that acceleration techniques 3-5 are not required when $n$ is small. For instance, *Chen et al.* [35] did not use techniques 3-5 for MNIST and CIFAR10 dataset. Although the attack success rate is compared to the success rate of the C&W attack, the number of required queries are large for gradient estimation despite the proposed acceleration techniques.

**Square Attack**

*Andriushchenko et al.* [26] proposed square attack (SA), a query-efficient attack on both $L_\infty$ and $L_2$-bounded adversarial perturbations. The SA is based on a random search (RS), an iterative technique in DFO methods. Therefore, the SA is a gradient-free attack [39] and it is resistant to gradient masking. SA improves the query efficiency and success rate by employing RS and a task-specific sampling distribution. In some cases, SA even competes with white-box attacks' performance.

Compared to an untargeted attack, SA is to find a $L_p$-norm bounded adversarial example by solving the following box-constrained optimization problem:

$$\min_{x'} L(f(x', y)) \qquad (4.4.3)$$

subject to $\| x' - x \|_p \leq \epsilon$ where $p \in \{2, \infty\}, x' \in [0,1]^n$, and $L(f(x', y)) = [f(x')]_y - \max_{j \neq y}[f(x')]_j$ is a margin-based loss function measuring the level of confidence that the model $f$ labels an adversarial input $x'$ the ground-truth class $y$ over other classes.

In a targeted attack with target class $t$, an attacker could simply minimize the loss function

$$L(f(x', t)) = \max_{j \neq t}[f(x')]_j - [f(x')]_t. \qquad (4.4.4)$$

However, a cross-entropy loss of the target class $t$

$$L(f(x', t)) = -f_t(x') + \log\left(\sum_{i=1}^{K} e^{f_i(x)}\right) \qquad (4.4.5)$$

is used to make the targeted attack more query efficient.

This optimization problem can be solved by using the classical RS method. In general, the RS method can be described as follows. Given an objective function $g$ to minimize, a starting point $x_0$, and a sampling distribution $D$, the RS algorithm outputs an estimated solution $x_{out}$ after $N$ iterations. For each iteration $i$, RS algorithm starts with a random update $\delta \sim D(x_i)$ and then adds this update to the current iteration; i.e., $x_{new} = x_i + \delta$. If the objective function $g_{new}$ evaluated at $x_{new}$ is less than the best $g^*$ obtained so far, we update $x_i$ and $g^*$. That is, if $g_{new} < g^*$, then $x_i = x_{new}$ and $g^* = g_{new}$. Otherwise, the update $\delta$ is discarded [40]. This process stops after a user specified number of iterations has reached. Thus, using RS to optimize $g$ does not rely on the gradient of $g$.

Since *Moon et al.* [41] showed that when one runs the white-box PGD attacks until convergence on CIFAR10 and ImageNet dataset, the successful adversarial perturbation $\delta$ is mainly found on vertices (corners) of $L_\infty$-norm ball [26]. Based on this observation, the SA attack only searches over the boundaries of the $L_\infty$ or $L_2$-ball to obtain the potential adversarial perturbation of the current iterate $\delta_i$. Hence, the perturbation for each pixel (before projection) is either $-2\epsilon, 0$, or $2\epsilon$ for the $L_\infty$-norm attack, where the perturbation of zero means no perturbation for a given pixel. This critical design makes the SA mechanism different from the standard RS algorithm, and it significantly improves the query efficiency.

Another key difference is that at each step the SA attack limits the modification of the current iterate $x'_i$ by updating only a small fraction (a square of side length $v_i$ for each color channel) of the neighboring pixels of $x_i'$. This helps reduce the dimension of the search space, especially when the input space $[0, 1]^n$ is high-dimensional, e.g., $n = 290 \times 290 \times 3$ in the ImageNet dataset [42].

In summary, the SA via random search consists of the following main steps for each iteration $i$ [40]:

1. Find the side length $v_i$ of the square by determining the closest positive integer to $\sqrt{p * w^2}$, where $p \in [0, 1]$ is the percentage of pixels of the original image $x$ that can be perturbed and $w$ is the width of an image. $p$ gradually decreases with the iterations but $v_i \geq 3$ for the $L_2$-norm SA. This mimics the step size reduction in gradient-based optimization. In gradient-based optimization, we begin with initial large learning rates to quickly shrink the loss value [40].

2. Find the location of the square with side length $v_i$ for each color channel by uniformly random selection. The square denotes the set of pixels that can be modified.

3. Uniformly assign all pixels' values in the square to either $-2 \times \epsilon$ or $+2 \times \epsilon$ for each color channel $c$.

4. Add the square perturbation generated in step 3 to the current iterate $x_i$ to obtain the new point $x_{i+1}$.

5. Project $x_{i+1}$ onto the intersection of $[0, 1]^n$ and the $L_\infty$-norm ball of radius $\epsilon$ to obtain $x_{new}$.

6. If the new point $x_{new}$ attains a lower loss than the best loss so far, the change is accepted, and the best loss is updated. Otherwise, it is discarded.

The iteration continues until the adversarial example is found. If interested, see [26].

## 4.4.3 Decision-based Attack

The decision-based attack uses only the label of an ML output for generating adversarial attacks, and this makes it easily applicable to real-world ML attacks. In this section, we discuss three popular decision-based black-box attacks, a boundary attack [43], a HopSkipJump attack [44], and a spatial-transformation attack [4].

**Boundary Attack**

Relying neither on training data nor on the assumption of transferability, the boundary attack uses a simple rejection sampling algorithm with a constrained independent and identically distributed Gaussian distribution as a proposed distribution and a dynamic step-size adjustment inspired by Trust Region methods to generate minimal perturbation adversarial samples. The boundary attack algorithm is given as follows. First, a data point is sampled randomly from either a maximum entropy distribution (for an untargeted attack) or a set of data points belonging to the target class (for a targeted attack). The selected data point serves as a starting point. At each step of the algorithm, a random perturbation is drawn from a proposed distribution such that the perturbed data still lies within the input domain and the difference between the perturbed image and the original input is within the specified maximum allowable perturbation $\epsilon$. Newly perturbed data is used as a new starting point if it is misclassified for an untargeted attack (or misclassified as the target class for a targeted attack). The process continues until the maximum number of steps is reached.

The boundary attack is conceptually simple, requires little hyperparameter tuning, and performs as well as the state-of-the-art gradient attacks (such as the C&W attack) in both targeted and untargeted computer vision scenarios without algorithm knowledge [43]. Furthermore, it is robust against common deceptions such as gradient obfuscation or masking, intrinsic stochasticity, or adversarial training. However, the boundary attack has two main drawbacks. First, the number of queries for generating an adversarial sample is large, making it impractical for real-world applications [44]. Instead of a rejection sampling algorithm, Metropolis-Hastings's sampling may be a better option since it does not simply discard the rejected sample. Secondly, it only considers $L_2$-norm.

**HopSkipJump Attack**

Conversely, the HopSkipJump attack is a family of query-efficient algorithms that generates both targeted and untargeted adversarial examples for both $L_2$ and $L_\infty$-norm distances. Furthermore, the HopSkipJump attack is more query efficient than the Boundary attack [45], and it is a hyperparameter-free iterative algorithm. The HopSkipJump attack is defined as follows.

> **Definition 13 (HopSkipJump Attack by *Chen et al.* 2020):** A HopSkipJump attack is a decision-based attack that generates the adversarial example $x'$ by solving the following optimization problem:
> $$\min_{x'} \| x' - x \|_p, \tag{4.4.6}$$
> subject to the constraint $\phi_x(x') = 1$, where $p = \{2, \infty\}$,
> $$\phi_x(x') = sign(S_{x^*}(x')) = \begin{cases} 1, & S_{x^*}(x') > 1 \\ -1, & \text{otherwise,} \end{cases} \tag{4.4.7}$$
> and
> $$S_{x^*}(x') = \begin{cases} [f(x')]_t - \max_{j \neq t}[f(x')]_j, \text{target on } t \\ \max_{j \neq y}[f(x')]_y - [f(x')]_y, \text{untargeted.} \end{cases} \tag{4.4.8}$$

Note that $S_{x^*}$ is similar to (4.3.18) with $\kappa = 0$. Let us discuss HopSkipJump $L_2$ based target attack in detail. Interested readers can consult [44] for untargeted attack or $L_2$ based attacks.

The HopSkipJump $L_2$ based target attack is an iterative algorithm consisting mainly an initialization step (step 1) and three iterative steps (steps 2-4 are repeated until the maximum number of iterations specified by an attacker is reached.):

1. Select a random data point $\tilde{x}_0$ from a target class (similar to the initialization step of a boundary attack).

2. Approach the decision boundary/hyperplane by using a binary search algorithm. That is, move $\tilde{x}_t$ towards the input $x$ and the decision hyperplane by interpolating between the input $x$ and $\tilde{x}_t$ to get a new data point $x_t$ such that $\phi_x(x_t) = 1$; i.e.,
$$x_t = \alpha_t x + (1 - \alpha_t)\tilde{x}_t, \tag{4.4.9}$$
where $\alpha_t$ is obtained via a binary search between 0 and 1 and stops at $x_t$ such that $\phi_x(x_t) = 1$.

3. Estimate the gradient-direction using the method similar to FGSM, see (4.3.8). That is,
$$v_t = \frac{\widehat{\nabla S}(x_t, \delta_t, B_t)}{\|\widehat{\nabla S}(x_t, \delta_t)\|_2} \tag{4.4.10}$$
where $\delta_t = \frac{\|x_t - x\|_2}{n}$ depends on the distance between $x_t$ and $x$ and the image size $n$, $\{u_b\}_{b=1}^{B_t}$ is a set of independent and identically distributed uniform random noise, $B_t = B_0 \sqrt{t}$ is the batch size of $\{u_b\}_{b=1}^{B_t}$, the initial batch size $B_0$ is set to 100 in [53], and
$$\widehat{\nabla S}(x_t, \delta_t, B_t) = \frac{1}{B_t - 1} \sum_{b=1}^{B_t} [\phi_x(x_t + \delta_t u_b) - \frac{1}{B_t} \sum_{b=1}^{B_t} \phi_x(x_t + \delta_t u_b)] u_b \tag{4.4.11}$$

4. Estimate the step size $\xi_t$ by geometric progression and then update the sample point $\tilde{x}_t$. That is,

starting with $\xi_t = \frac{||x_t - x||_2}{\sqrt{t}}$, the step size is reduced by half until $\phi_x(\tilde{x}_t) = 1$, where

$$\tilde{x}_t = x_t + \xi_t v_t = x_t + \xi_t \frac{\widehat{\nabla S}(x_t, \delta_t, B_t)}{||\widehat{\nabla S}(x_t, \delta_t)||_2}. \qquad (4.4.12)$$

**Spatial Transformation Attack**

In contrast to the boundary attack and the HopSkipJump attack that based on $L_p$ bounded adversarial perturbations, the spatial transformation attack [4] is a black-box "Semantic" adversarial attack that reveals a small random transformation, such as a small rotation on an image, can fool a state-of-the-art image classifier easily. This attack really questions the robustness of these state-of-the-art image classifiers. Furthermore, the primary disadvantage of the ZOO attack is the need to probe the classifiers thousands of times before the adversarial examples are found [27]. This is not a case for spatial transformation attacks. *Engstrom et al.* [4] showed that worst-of-10 ($k = 10$) is able to reduce the model accuracy significant with just 10 queries. Basically, it rotates or translates a natural image slightly to cause misclassification. This attack reveals the vulnerability of the current state-of-the-art ML models.

> **Definition 14 (Spatial Transformation Attack):** Let $x \in R^n$ be a legitimate input image that is correctly classified as class $y$ by an image classifier. A spatial transformation attack generates an adversarial example $x' \in R^n$ by finding a mapping $\alpha: R^n \to R^n$ such that the adversarial example $\alpha(x) = x'$, where the relationship between $x'$ and $x$ is as follows:
>
> Each pixel $(a, b, c)$ in $x$ is translated $(\delta a, \delta b, 0)$ pixels to the right and rotate $\gamma$ degrees counterclockwise around the origin to obtain the pixel $(a', b', c)$ in $x'$, where $a' = a * cos\, cos\,(\gamma) - b * sin\, sin\,(\gamma) + \delta a$ and $b' = a * sin\, sin\,(\gamma) + b * cos\, cos\,(\gamma) + \delta b$.
>
> To find the mapping $\alpha$, an attacker needs to solve the following optimization problem:
>
> $$\max_{a,b,c} L(\alpha(x), y), \qquad (4.4.13)$$
>
> where $L$ is the loss function.

*Engstrom et al.* [4] proposed three different methods to solve (4.4.13): (1) first-order method to maximize the loss function, (2) grid search on a discretized parameter space, and (3) worst-of-$k$ (i.e., randomly sample $k$ different set of attack parameters and choose one that model performs worst). The first order method requires gradient of the loss function which is not possible in black box setting. Even if it is possible, *Engstrom et al.* [4] shows that the first-order method performs worst due to the non-concavity of loss function for spatial perturbation.

## 4.5 DATA POISONING ATTACKS

While adversarial attacks cannot change the training process of a model and can only modify the test instance, the data poisoning attacks, on the contrary, can manipulate the training process. Specifically, in data poisoning attacks, attackers aim to manipulate the training data (e.g., poisoning features, flipping labels, manipulating the model configuration settings, and altering the model weights) in order to influence the learning model. It is assumed that attackers have the capability to contribute to the training data or have control over the training data itself. The main objective of injecting poison data is to influence the model's learning outcome.

Recent studies on adversarial ML have demonstrated particular interest in data poisoning attack settings. This section discusses few data poisoning attack models. We start with briefly going over label flipping attacks in subsection 4.5.1 and then focus on clean label data poisoning attacks in subsection 4.5.2 since they are stealthy. To the end, we introduce backdoor attacks.

## 4.5.1 Label Flipping Attacks

A simple and effective way to attack a training process is to simply flip labels of some training instances. This type of data poisoning attacks is called label flipping attacks. Mislabeling can be done easily in crowdsourcing, where an attacker is one of the annotators, for example. In this subsection, we discuss some common label flip attacks against ML models. These label flip attacks could be either model-independent or model-dependent. For instance, a random label flipping attack is model-independent. It simply selects a subset of training instances and flips their labels. The attacker does not need to have any knowledge of the underlying target ML model. Even though this random strategy looks simple, it is capable of reducing classification accuracy significantly depending on the type of dataset under attack, the training set size, and the portion of the training labels that are flipped. The random label flip attack can be mathematically defined as follows.

> **Definition 15 (Random Label Flipping Attack):** Given a training set $\{x_i, y_i\}_{i=1}^n$, where $x_i \in X$ and $y_i \in \{-1, 1\}$, it is called a random label flipping attack if an attacker with the ability to change $p \in (0, 1)$ fraction of the training label can randomly select $\lfloor np \rfloor$ training labels and flip the labels.

This random label flipping attack can further be divided into two groups: targeted and untargeted. In an untargeted random label flipping attack, an attacker may select some instances from class 1 to misclassify as class $-1$ and some instances from class $-1$ to misclassify as class 1. In contrast, an attacker misclassifies one class as another consistently in a targeted random label flipping attack. The targeted random label flipping attack is more severe compared to the untargeted one as the targeted attack consistently misleads the learning algorithm to classify a specific class of instances as another specific class.

Rather than random label flipping, an attacker can also utilize label flip attacks that are model-dependent. For instance, we show that SVM constructs a decision hyperplane using only the support vectors in subsection 4.2.2. Existing studies presented a few label flipping attacks based on this characteristic. For example, Farfirst attack [46] is one of such label flipping attacks, where a training instance far away from the margin of an SVM is flipped. This attack effectively changes many non-support vector training instances (training instances that are far from the margin and correctly classified by an untainted SVM) to support vectors, and significantly alter the decision boundary consequently.

Formally, an optimal label flipping attack can be considered as a bi-level optimization problem defined as follows.

> **Definition 16 (Optimal Label Flipping Attack):** Suppose that training set $D = \{x_i, y_i\}_{i=1}^n$ and a test set $T = \{\tilde{x}_i, \tilde{y}_i\}_{i=1}^m$, where $x_i \in X, \tilde{x}_i \in X, y_i \in \{-1, 1\}$, and $\tilde{y}_i \in \{-1, 1\}$. Let $l$ be the number of training labels that an attacker has the ability to modify, and let $I \in \{-1, 1\}^n$ be an indicator vector such that its $i^{th}$ element $I_i$ equals to $-1$ if the $i^{th}$ training label is flipped and 1 otherwise. Then, $|I| = l$ and the poisoned training set $D' = \{x_i, y_i'\}_{i=1}^n$, where $y_i' = y_i I_i$. It is called an optimal label flipping attack if an attacker can find an optimal $I$ by solving the following optimization problem:
>
> $$O(\min_{\Theta} L(D', \Theta), T), \quad (4.5.1)$$
>
> where $O$ is an objective function specified by the attacker. A common objective function is to maximize the test error rate, in which case $O(\Theta^*, T) = \frac{1}{m}\sum_{i=1}^m J[C(\tilde{x}_i) = \tilde{y}_i]$, where $J$ is an indicator function such that
>
> $$J[C(\tilde{x}_i) = \tilde{y}_i] = \begin{cases} 1, & C(\tilde{x}_i) = \tilde{y}_i \\ 0, & otherwise \end{cases}. \quad (4.5.2)$$

However, solving such a bi-level optimization problem is NP-hard. Here, we introduce a simple greedy algorithm that is suboptimal but tractable. At iteration $t = 1$, an attacker first flips the label of the first training instance to obtain a poisoned training set $D'$. Then, the attacker trains an ML model using $D'$ and evaluates the performance $p_1$ of the poisoned model. Next, the attacker flips the label of the second training instance while keeping other labels untainted. Similarly, the attacker trains the ML model using the poisoned dataset and evaluates its performance $p_2$ on the test set. The attacker repeats this process for each training label to obtain $\{p_1, p_2, ..., p_n\}$. Now, the attacker can determine which training label is actually flipped by simply finding the training label corresponding to the worst performance. After that label is flipped, the attacker can select the next training label to flip out of the remaining $n - 1$ training labels by following the same process as iteration 1. For each iteration, one additional training label is flipped. The process stops until $l$ training labels have been flipped.

## 4.5.2 Clean Label Data Poisoning Attack

Security of DL networks can be degraded by the emergence of clean label data poisoning attacks. This is achieved by injecting legitimately labeled poison samples into a training set. Although poison samples seem normal to a user, they indeed comprise illegitimate characteristics to trigger a targeted misclassification during the inference process.

In this subsection, an attacker deploys "clean-labels" data poisoning attacks without the need to control the labeling process. Furthermore, such adversarial attacks are often "targeted" such that they allow for controlling the classifier's behavior on a particular test instance while maintaining the overall performance of the classifier [47]. For instance, an attacker can insert a seemingly legitimate image, i.e., a correctly labeled image, into a training dataset for a facial identification engine and thus control the chosen person's identity during the test time. This attack is severe since the poison instance can be placed online and awaits to be scraped by a bot for data collection. The poison instance is then labeled correctly by a trusted annotator but still able to subvert the learning algorithm.

Here, we introduce two clean label data poisoning attacks against both transfer learning and an end-to-end training.

**Feature Collision Attack**

The first one is the feature collision attack introduced by *Shafahi et al.* [47] who assumed that an attacker has

- no knowledge of the training data,
- no ability to influence the labeling process,
- no ability to modify the target instance during inference, and
- knowledge of the feature extractor's architecture and its parameters.

An attacker tries to find a poison instance that collides with a given target instance in a feature space while maintaining its indistinguishability with a base instance from class $c$ other than target class $t \neq c$. Hence, the generated poison instance looks like a base instance and an annotator labels it as an instance from class $c$. However, that poison instance is close to the target instance in a feature space, and the ML model is likely classifying it as an instance from class $t$. This causes the targeted misclassification and only one such poison instance is needed to poison a transfer learning model. That is why this attack is sometimes called a one-shot attack as well, and its formal definition is shown as follows.

**Definition 17 (Feature Collision Attack by *Shafahi et al.* 2018):** Given a feature extractor $f$, a target instance $x_t$ from class $t$, and a base instance $x_b$ that belongs to a targeted class $b$ such that $b \neq t$, it is called a feature collision attack if an attacker finds a poison instance $x_p$ as follows:

$$x_p = \underset{x}{\mathrm{argmin}} \|f(x) - f(x_t)\|_2^2 + \beta \|x - x_b\|_2^2, \qquad (4.5.3)$$

where $\beta$ is a similarity parameter that indicates the importance of the first component $\|f(x) - f(x_t)\|_2^2$ (the first component measures the similarity of the poison instance and the target instance in the feature space created by $f$) over the second component $\|x - x_b\|_2^2$, which measures the similarity of the poison instance and the base instance in the input space.

The base instance can be selected randomly from any classes other than the target class. However, some base instances may be easier for an attacker to find a poison instance than others. The coefficient $\beta$ must be tuned by the attacker in order for them to make the poison instance seem indistinguishable from a base instance. *Shafahi et al.* [47] solved the optimization problem by using a forward-backward-splitting iterative procedure [48].

This attack has a remarkable attack success rate (e.g., 100% in one experiment presented in [47]) against transfer learning. Nevertheless, we want to point out that such an impressive attack success rate reported in [47] is due to the overfitting of the victim model to the poison instance. The data poisoning attack success rate drops significantly on an end-to-end training and in black-box settings, and *Shafahi et al.* [47] proposed to use a watermark and multiple poison instances to increase the attack success rate on an end-to-end training. However, one obvious drawback is that the pattern of the target instance shows up in the poison instances sometimes.

**Convex Polytope Attack and Bullseye Polytope Attack**

Although in many ML-based attacks, an attacker is required to obtain full access to the network architecture and weight of a model (e.g., a feature collision attack), such a privilege is likely protected in commercialized products such as perception systems in smart cars. In this subsection, we introduce a transferable and scalable clean label targeted data poisoning attack against transfer learning and an end-to-end training in a black-box setting, where attackers do not have access to the target model. *Aghakhani et al.* [49] proposed such a clean label data poisoning attack model, named the bullseye polytope attack, to create poison instances that are similar to a given target instance in a feature space. The bullseye polytope attack achieves a higher attack success rate compared to a feature collision attack in a black-box setting and increases the speed and success rate of the targeted data poisoning attack on the end-to-end training compared to the convex polytope attack.

The convex polytope attack [50] is very similar to the bullseye polytope attack; both of them are the extension of a feature collision attack into a black-box setting. Let us define the convex polytope attack formally as follows before we discuss the bullseye polytope attack, a scalable alternative to the convex polytope attack.

**Definition 18 (Convex Polytope Attack by *Zhu et al.* 2019):** Given a set of feature extractors $\{f^i\}_{i=1}^m$, a target instance $x_t$ from class $t$ and a set of base instances $\{x_b^j\}_{j=1}^k$ that belongs to class $b$ such that $b \neq t$, it is called a convex polytope attack if an attacker finds a set of poison instances $\{x_p^j\}_{j=1}^k$ by solving the following optimization problem:

$$\min_{\substack{\{x_p^j\}_{j=1}^k \\ \{c^i\}_{i=1}^m}} \frac{1}{2} \sum_{i=1}^m \left( \frac{\|f^i(x_t) - \sum_{j=1}^k c_j^i f^i(x_p^{(j)})\|^2}{\|f^i(x_t)\|^2} \right), \qquad (4.5.4)$$

subject to the constraints $\sum_{j=1}^k c_j^i = 1$, $c_j^i \geq 0$, $\forall i = 1, 2, \ldots, m$, $\forall j = 1, 2, \ldots, k$, and $\|x_p^j - x_b^j\|_\infty \leq \epsilon$, $\forall j = 1, 2, \ldots, k$, where $c^i \in R^k$ is a coefficient vector consisting of elements $c_j^i$.

Without the knowledge of a victim model, an attacker can utilize the attack transferability to attack the victim model by substituting feature extractors $\{f^i\}_{i=1}^m$ (similar to transfer attacks in section 4.4.1). This is the main idea behind both the convex polytope attack and the bullseye polytope attack. The main difference between the two attacks is that there is no incentive to refine the target deeper inside the convex hull of poisons ("attack zone") for the convex polytope attack, whereas the bullseye polytope attack pushes the target into the center of convex hull by solving the optimization problem (4.5.5) instead. Consequently, the target is often near the convex hull boundary in the convex polytope attack. The optimization problem (4.5.4) requires finding a set of coefficient vectors $\{c^i\}_{i=1}^m$ in addition to find a set of poisons $\{x_p^j\}_{j=1}^k$ that minimizes its objective function. Hence, the convex polytope attack has a much higher computation cost than that of the bullseye polytope attack. As shown below, the bullseye polytope attack simply sets $c_j^i = \frac{1}{k}, \forall i = 1, 2, \ldots, m, \forall j = 1, 2, \ldots, k$, to assure that the target remains near the center of the poison samples.

---

**Definition 19 (Single-target Mode Bullseye Polytope Attack by *Aghakhani et al.* 2020):** Given a set of feature extractors $\{f^i\}_{i=1}^m$, a target instance $x_t$ from class $t$ and a set of base instances $\{x_b^j\}_{j=1}^k$ that belongs to class $b$ such that $b \neq t$, it is called a single-target mode bullseye polytope attack if an attacker finds a set of poison instances $\{x_p^j\}_{j=1}^k$ by solving the following optimization problem:

$$\min_{\{x_p^j\}_{j=1}^k} \frac{1}{2} \sum_{i=1}^m \left( \frac{\|f^i(x_t) - \frac{1}{k} \sum_{j=1}^k f^i(x_p^{(j)})\|^2}{\|f^i(x_t)\|^2} \right), \quad (4.5.5)$$

subject to the constraints $\|x_p^j - x_b^j\|_\infty \leq \epsilon, \forall j = 1, 2, \ldots, k$, where $\epsilon$ determines the maximum allowed perturbation.

---

The bullseye polytope attack can further be extended to a more sophisticated target mode (called the multi-target mode) by constructing a set of poison images that targets multiple instances enclosing the same object from different angles or lighting conditions. Specifically, if $n_t$ target instances of the same object is used in a multi-target mode attack, $f^i(x_t)$ in (4.5.5) is replaced by the average target feature vectors $\sum_{v=1}^{n_t} f_v^i(x_t)$. Such a model extension can remarkably enhance the transferability of the attack to unseen instances comprising the same object without having to use additional poison samples [49].

## 4.5.3 Backdoor Attack

In this last subsection, we introduce backdoor attacks, a type of data poisoning attacks/causative attacks that contains a backdoor trigger (an adversary's embedded pattern). Specifically, we consider backdoor attacks on an outsourcing scenario [51], a transfer learning scenario [51], and a federated learning scenario [52]. Here are the three attack scenarios:

- *Outsourced training*: In this attack scenario, a user aims to train the parameters Θ of a network $f_\Theta$ by using a training dataset. For this purpose, the user sends the model description to the trainer who will then return trained parameters Θ′. The user herein verifies the trained model's parameters by checking the classification accuracy on a "held-out" validation dataset, $D_{valid}$, as the trainer cannot be fully trusted. The model will be accepted only if its accuracy fulfills a target accuracy, $a^*$, on the validation set. Building upon the above setting, a malicious trainer will return an illegitimate backdoored model $f_{\Theta_{adv}}$ to the user such that $f_{\Theta_{adv}}$ has met the target accuracy requirement while $f_{\Theta_{adv}}$ misclassifies the backdoored instances.

- *Transfer learning*: In this attack scenario, a user unintentionally downloads an illegitimately trained model $f_{\Theta_{adv}}$ and the corresponding training and validation set $D_{valid}$ from an online repository. Then, the user develops a transfer learning model $f_{\Theta_{adv},TL}$ based on it. The main attacker's goal is similar to his/her goal in the adversarial outsourced training scenario. However, both $f_{\Theta_{adv}}$ and $f_{\Theta_{adv},TL}$ must have the reasonably high accuracy on $D_{valid}$ and the user's private validation set for the new domain of application.

- *Federated learning*: This attack scenario is especially vulnerable to a backdoor attack since, by design, the central server has no access to the participants' private local data and training. A malicious participant or a compromised participant can manipulate the joint model by providing a malicious interned update to the central server based on the techniques such as contrain-and-scale [52]. As a result, the joint model would behave in accordance with the attacker's goal as long as the input encloses the backdoor features while the model can maintain the classification accuracy on the clean test instances.

An attacker can generate an illegitimate trained DL network, also known as a backdoored neural network (BadNet), for both targeted or untargeted misclassification on backdoor instances by modifying the training procedure, such as injecting poisoned instances with backdoor trigger superimposed and label altered, manipulating the model configuration settings, or altering model parameters directly. Subsequently, the malicious learning network behaves wrongly on backdoor instances whereas has high classification accuracy on the user's private validation set.

Although various detection mechanisms have been elaborated on adversarial backdoors detection, e.g., using statistical differences in latent representations between clean input data and attacker's input samples in a poisoned model, we introduce an adversarial backdoor embedding attack [53] that can bypass several detection mechanisms (e.g., Feature Pruning [54] and Dataset Filtering by Spectral Signatures [55]) simultaneously in this section. Notably, *Shokri et al.* [53] developed an adversarial training mechanism, which creates a poisoned model that can mitigates either a specific defense or multiple defenses at the same time by utilizing an objective function that penalizing the difference in latent representations between the clean inputs and backdoored inputs.

The proposed mechanism maximizes the indistinguishability of both clean and poisoned data's hidden representations by utilizing the idea similar to a generative adversarial network. The poisoned model decouples the adversarial inputs comprising the backdoor triggers from clean data by developing a discriminative network. The conventional backdoor detection mechanisms may succeed in recognizing the backdoor triggers only if the attacker naively trains the poisoned model in a way that leaves a remarkable difference in the distribution of latent representations between the clean data and backdoor instances. Therefore, a knowledgeable attacker can make the model more robust against existing backdoor detection mechanisms by minimizing the difference in latent representations of two. As a result, the attacker can attain high accuracy of classification by the poisoned model (the behavior of the poisoned model remains unmodified on clean data), while fooling the backdoor detection mechanism.

## 4.6 CONCLUSIONS

As ML systems have been dramatically integrated into a broad range of decision-making-sensitive applications for the past years, adversarial attacks and data poisoning attacks have posed a considerable threat against these systems. For this reason, this chapter focuses on the two important areas of ML security: adversarial attacks and data poisoning attacks. Specifically, this chapter has studied the technical aspects of these two types of security attacks. It has comprehensively described, discussed, and scrutinized the adversarial attacks and the data poisoning attacks with regard to their applicability requirements and adversarial capabilities. The main goal of this chapter is to help the research community gain insights and implications of existing adversarial attacks and data poisoning attacks, as well as to increase the awareness of potential adversarial threats when ones develop learning algorithms and apply ML methods to various applications in the future.

Researchers have developed many adversarial attack and data poisoning attack approaches over the years, but we are not able to cover all of them due to the space limit. For instance, *Biggio et al.* [56] proposed a set of poisoning attacks against SVMs, where attackers aim to increase the SVM's test error in 2012. In this study, *Biggio et al.* also demonstrated that an attacker could predict the SVM's decision function change, which can then be used to construct malicious training data. Furthermore, *Li et al.* [57] developed a spatial-transformation robust backdoor attack in 2020. Moreover, many defense mechanisms have been proposed as well. If interested, please see [28, 50, 54] [57-64].

# ACKNOWLEDGMENT

We acknowledge the National Science Foundation (NSF) for partially sponsoring the work under grants #1633978, #1620871, #1620862, and #1636622, and BBN/GPO project #1936 through an NSF/CNS grant. We also thank the Florida Center for Cybersecurity (Cyber Florida) for a seed grant. The views and conclusions contained herein are those of the authors and should not be interpreted as necessarily representing the official policies, either expressed or implied of NSF.